\documentclass[letterpaper, 10 pt, journal, twoside]{IEEEtran}

\IEEEoverridecommandlockouts                              % This command is only needed if 
\usepackage{graphics} % for pdf, bitmapped graphics files
\usepackage{epsfig} % for postscript graphics files
\usepackage{amsmath} % assumes amsmath package installed
\usepackage{amssymb}  % assumes amsmath package installed
\usepackage{hyperref}
\usepackage{threeparttable}
\usepackage{siunitx}
\hypersetup{
	colorlinks=true,
	linkcolor=blue,
	filecolor=magenta,      
	urlcolor=blue,
}

\usepackage{xcolor}
\title{
TH\"OR: Human-Robot Navigation Data Collection \\and Accurate Motion Trajectories Dataset
}

\author{Andrey Rudenko$^{1,2}$, Tomasz P. Kucner$^{2}$, Chittaranjan S. Swaminathan$^{2}$, Ravi T. Chadalavada$^{2}$,\\ Kai O. Arras$^{1}$ and Achim J. Lilienthal$^{2}$%
\thanks{Manuscript received: September 10, 2019; Accepted November 28, 2019.}%Use only for final RAL version
\thanks{This paper was recommended for publication by Editor D. Lee upon evaluation of the Associate Editor and Reviewers' comments. This work has been partly funded from the European Union's Horizon 2020 research and innovation programme under grant agreement No 732737 (ILIAD) and by the Swedish Knowledge Foundation under contract number 20140220 (AIR).}
\thanks{$^{1}$A. Rudenko and K.O. Arras are with Bosch Corporate Research, Stuttgart, Germany {\tt\footnotesize \{andrey.rudenko, kaioliver.arras\}@de.bosch.com}}%
\thanks{$^{2}$A. Rudenko, T. Kucner, C. Swaminathan, R. Chadalavada and A. Lilienthal are with the Mobile Robotics and Olfaction Lab, \"Orebro University, Sweden {\tt\footnotesize \{Tomasz.Kucner, Chittaranjan.Swaminathan, Ravi.Chadalavada, Achim.Lilienthal\}@oru.se}}
\thanks{Digital Object Identifier (DOI): see top of this page.}
}

\begin{document}

\maketitle
%\thispagestyle{empty}
%\pagestyle{empty}

% Paper headers
\markboth{IEEE Robotics and Automation Letters. Preprint Version. Accepted November 2019}
{Rudenko \MakeLowercase{\textit{et al.}}: TH\"OR: Human-Robot Navigation Data Collection and Accurate Motion Trajectories Dataset} 
% Use only for final RAL version

%%%%%%%%%%%%%%%%%%%%%%%%%%%%%%%%%%%%%%%%%%%%%%%%%%%%%%%%%%%%%%%%%%%%%%%%%%%%%%%%
\begin{abstract}

%Robots increasingly often have to co-exist with us in shared environments, such as the airports, shopping centers and hospitals. Safe and efficient operation in such scenarios poses new challenges to human-aware motion and task planners. In particular, the robots need to read the intention of surrounding people from non-verbal cues, predict motion trajectories and learn typical mobility patterns of their users. In this paper we present a large-scale experimental study of human navigation in shared environments and describe a new dataset of indoor Trajectories of Humans, recorded in the \"ORebro university (TH\"OR).

%Robots increasingly often have to co-exist with us in shared environments, such as the airports, shopping centers and hospitals. Safe and efficient operation in such scenarios poses new challenges to human-aware motion and task planners. In particular, the robots need to read the intention of surrounding people from non-verbal cues, predict motion trajectories and learn typical mobility patterns of their users. In this paper we present a large-scale experimental study of human navigation in shared environments and describe a new dataset of indoor Trajectories of Humans, recorded in the \"ORebro university (TH\"OR). In a quantitative analysis using several metrics we show that the TH\"OR dataset contains higher variety of motion behavior, less noise and higher tracking duration than three popular datasets.

Understanding human behavior is key for robots and intelligent systems that share a space with people. Accordingly, research that enables such systems to perceive, track, learn and predict human behavior as well as to plan and interact with humans has received increasing attention over the last years. The availability of large human motion datasets that contain relevant levels of difficulty is fundamental to this research. Existing datasets are often limited in terms of information content, annotation quality or variability of human behavior. In this paper, we present TH\"OR, a new dataset with human motion trajectory and eye gaze data collected in an indoor environment with accurate ground truth for position, head orientation, gaze direction, social grouping, obstacles map and goal coordinates. TH\"OR also contains sensor data collected by a 3D lidar and involves a mobile robot navigating the space. We propose a set of metrics to quantitatively analyze motion trajectory datasets such as the average tracking duration, ground truth noise, curvature and speed variation of the trajectories. In comparison to prior art, our dataset has a larger variety in human motion behavior, is less noisy, and contains annotations at higher frequencies.
\end{abstract}

% Keywords appear just beneath the abstract. Use only for final RAL version. 

%%%%%%%%%%%%%%%%%%%%%%%%%%%%%%%%%%%%%%%%%%%%%%%%%%%%%%%%%%%%%%%%%%%%%%%%%%%%%%%%
\section{Introduction}
\label{sec:introduction}

\IEEEPARstart{U}{nderstanding} human behavior has been the subject of research for autonomous intelligent systems across many domains, from automated driving and mobile robotics to intelligent video surveillance systems and motion simulation. %For safe and efficient operation in shared spaces such systems should be human-aware, understand and respect social rules.
Human motion trajectories are a valuable learning and validation resource for a variety of tasks in these domains. For instance, they can be used for learning safe and efficient human-aware navigation, predicting motion of people for improved interaction and service, inferring motion regularities and detecting anomalies in the environment. Particular attention towards trajectories, intentions and mobility patterns of people has considerably increased in the last decade \cite{rudenko2019human}.

Datasets of ground level human trajectories, typically used for learning and benchmarking, include the ETH \cite{pellegrini2009you}, Edinburgh \cite{majecka2009statistical} and the Stanford Drone \cite{robicquet2016learning} datasets, recorded outdoors, or the indoor ATC \cite{brscic2013person}, L-CAS \cite{yan2017online} or Central Station \cite{zhou2012understanding} datasets (see Table~\ref{tab:datasets}). While providing the basic input of motion trajectories, these datasets often lack relevant contextual information and the desired properties of data, e.g. the map of static obstacles, coordinates of goal locations, social information such as the grouping of agents, high variety in the recorded behaviors or long continuous tracking of each observed agent. Furthermore, most of the recordings are made outdoors, a robot is rarely present in the environment and the ground truth pose annotation, either automated or manual, is prone to artifacts and human errors.

\begin{figure}[t]
	\begin{center}
		\vspace{5pt}
		\includegraphics[width=0.99\columnwidth]{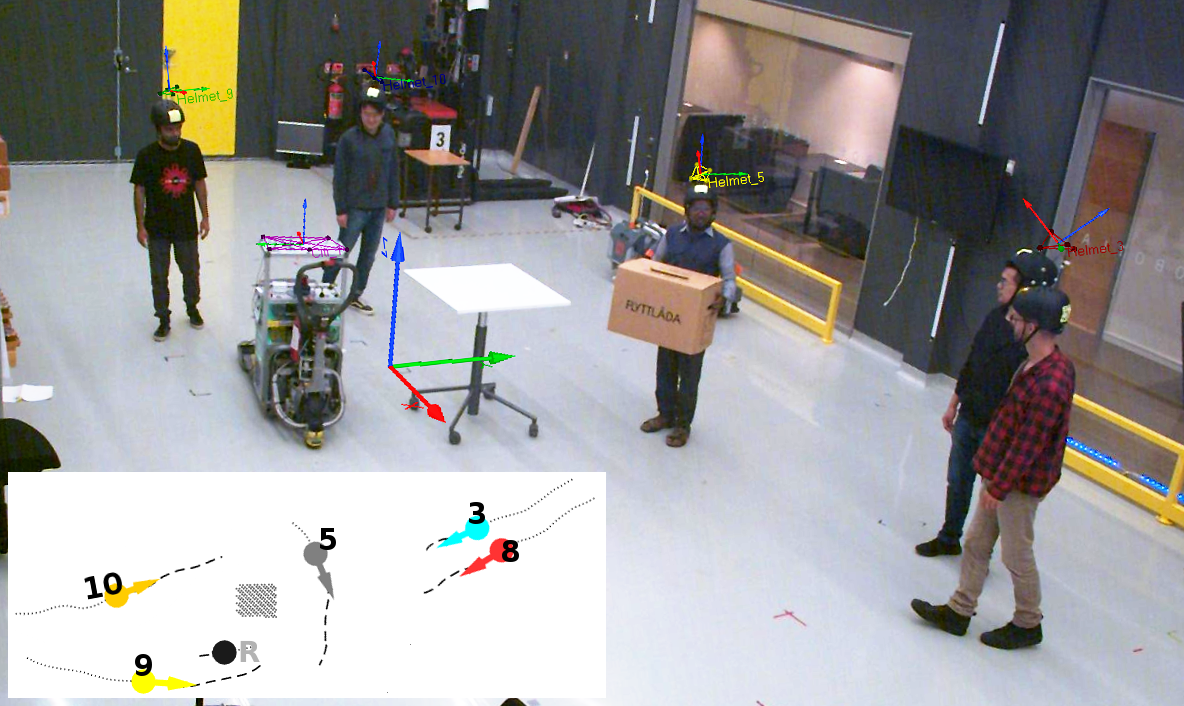}
	\end{center}
	\vspace{-8pt}
	\caption{Environment configuration. Participants, wearing tracking helmets, and the robot are moving towards their goals in a shared space, tracked by the Qualisys motion capture system (recorded motion in the bottom left corner).}
	\label{fig:cover}
	%\vspace{-6pt}
\end{figure}

\begin{table*}[t!]
	\center
	\begin{threeparttable}[t]
	\vspace{5pt}
	\footnotesize{
		\begin{tabular}{ | p{2.12cm}| p{1.0cm}| p{0.5cm}| p{1.05cm}| p{0.9cm}|p{1.32cm}|p{0.44cm}|p{1.12cm}|p{1.89cm}|p{1.12cm}|p{1.46cm}|}
			\hline
			\textbf{Dataset} & \textbf{Location} & \textbf{Map} & \textbf{Goal positions} & \textbf{Groups} & \textbf{Head orientation} & \textbf{Eye gaze} & \textbf{Robot in the scene} & \textbf{Sensors for pose estimation} & \textbf{Frequency} & \textbf{Annotation} \\ \hline
			ETH \cite{pellegrini2009you} & Outdoors & \checkmark & \checkmark & \checkmark &  &  &  & Camera & 2.5 Hz & Manual \\ \hline
			UCY \cite{lerner2007crowds} & Outdoors &  &  &  &  & \checkmark &  & Camera & Continuous & Manual \\ \hline
			VIRAT~\cite{oh2011large} & Outdoors & $\checkmark^*$ &  &  &  &  &  & Camera & 2,5,10 Hz & Manual \\ \hline
			KITTI \cite{Geiger2012CVPR} & Outdoors & \checkmark &  &  &  &  & \checkmark & Velodyne and several cameras & 10 Hz & Manual \\ \hline
			Edinburgh~\cite{majecka2009statistical} & Outdoors &  &  &  &  &  &  & Camera & 6-10 Hz (variable) & Automated \\ \hline
			Stanford~Drone~\cite{robicquet2016learning} & Outdoors & $\checkmark^*$ &  &  &  &  &  & Camera & 30 Hz & Manual \\ \hline
			Town~Center~\cite{benfold2011stable} & Outdoors & $\checkmark^*$ &  &  &  &  &  & Camera & 15 Hz & Manual \\ \hline
			ATC \cite{brscic2013person} & Indoors &  &  &  & \checkmark &  &  & Several 3D range sensors & 10-30 Hz (variable) & Automated \\ \hline
			Central station \cite{zhou2012understanding} & Indoors &  &  &  &  &  &  & Camera & 25 Hz & Automated \\ \hline
			L-CAS \cite{yan2017online} & Indoors &  &  & \checkmark &  &  & \checkmark & 3D LiDAR & 10 Hz & Manual \\ \hline
			KTH \cite{dondrup2015tracker} & Indoors &  &  &  &  &  & \checkmark & RGB-D, 2D laser scaner & 10-17 Hz (variable) & Automated \\ \hline
			TH\"OR & Indoors & \checkmark & \checkmark & \checkmark & \checkmark & \checkmark & \checkmark & Motion capture & 100 Hz & Ground truth \\ \hline
		\end{tabular}
		\begin{tablenotes}
			\item[*] Unsegmented camera image.
		\end{tablenotes}
	}
	\end{threeparttable}%
	\vspace{10pt}
	\caption{Datasets of human motion trajectories}
	\label{tab:datasets}
	\vspace{-10pt}
\end{table*}

In this paper we present a human-robot interaction procedure, designed to collect motion trajectories of people in a generic indoor social setting with extensive interaction between groups of people and a robot in a spacious environment with several obstacles. The locations of the obstacles and goal positions are set up to make navigation non-trivial and produce a rich variety of behaviors. The participants are tracked with a motion capture system; furthermore, several participants are wearing eye-tracking glasses. ``Tracking Human motion in the \"ORebro university'' (TH\"OR) dataset\footnote{Available at \url{http://thor.oru.se}}, which is released public and free for non-commercial purposes, contains over 60 minutes of human motion in 395k frames, recorded at 100 Hz, 2531k people detections and over 600 individual and group trajectories between multiple resting points. In addition to the video stream from one of the eye tracking headsets, the data includes 3D Lidar scans and a video recording from stationary sensors. We quantitatively analyze the dataset using several metrics, such as tracking duration, perception noise, curvature and speed variation of the trajectories, and compare it to popular state-of-the-art datasets of human trajectories. Our analysis shows that TH\"OR has more variety in recorded behavior, less noise, and high duration of continuous tracking.

The paper is organized as follows: in Sec.~\ref{sec:related_work} we review the related work and in Sec.~\ref{sec:experiment} detail the data collection procedure. In Sec.~\ref{sec:results} we describe the recorded data and analyze it quantitatively and qualitatively. Sec.~\ref{sec:conclusions} concludes the paper.

%Similarly to how challenging datasets promote and push forward innovations in computer vision \cite{oh2011large,schneider2013gcpr}, dedicated benchmarking is very important for human motion analysis.

\section{Related Work}
\label{sec:related_work}

Recordings of human trajectory motion and eye gaze are useful for a number of research areas and tasks both for machine learning and benchmarking. Examples include person and group tracking \cite{pellegrini2009you,lau2009tracking,linder2016multi}, human-aware motion planning \cite{Foka2010,Bai2015,palmieri2017kinodynamic,swaminathan2018down}, %, social science \cite{moussaid2010walking}
motion behavior learning \cite{okal2016learning}, human motion prediction \cite{chung2012incremental,Rudenko2018iros}, human-robot interaction \cite{lasota2017survey}, video surveillance \cite{alahi2016social} or collision risk assessment \cite{lo2019robust}. In addition to basic trajectory data, state-of-the-art methods for tracking or motion prediction, for instance, can also incorporate information about the environment, social grouping, head orientation or personal traits. For instance, Lau et al. \cite{lau2009tracking} estimate social grouping formations during tracking and Rudenko et al. \cite{Rudenko2018iros} use group affiliation as a contextual cue to predict future motion. Unhelkar et al. \cite{unhelkar2015human} use head orientation to disambiguate and recognize typical motion patterns that people are following. Bera et al. \cite{bera2017aggressive} and Ma et al. \cite{ma2016forecasting} learn personal traits to determine interaction parameters between several people. To enable such research in terms of training data and benchmarking requirements, a state-of-the-art dataset should include this information.

%Recordings of human motion are valuable for a number of tasks, such as designing human-aware motion planners \cite{Foka2010,Bai2015,lo2019robust} and investigating the principles of human motion \cite{moussaid2010walking}. In particular, human motion prediction research -- an important task in service robotics \cite{chung2012incremental,bayoumi2017learning,Rudenko2018iros}, automated industry \cite{lasota2017survey} and video surveillance \cite{alahi2016social} -- relies on recorded trajectories for training and benchmarking. State-of-the-art predictors are capable of inferring long-term, map-aware trajectories of people, considering also dynamic and semantic cues of the agents. For instance, Rudenko et al. \cite{Rudenko2018iros} present a group-aware predictor which combines multimodal uncertainty- and map-aware global motion prediction with local interaction modeling. Unhelkar et al. \cite{unhelkar2015human} use the head orientation to disambiguate the motion pattern, which the person is following. Bera et al. \cite{bera2017aggressive} and Ma et al. \cite{ma2016forecasting} learn personal traits to determine interaction parameters between several people. Appropriate training and benchmarking data is required to correctly realize and assess the capabilities of such advanced methods.

Human trajectory data is also used for learning long-term mobility patterns \cite{molina2018modelling}, such as the CLiFF maps \cite{kucner2017enabling}, to enable compliant flow-aware global motion planning and reasoning about long-term path hypotheses towards goals in distant map areas for which no observations are immediately available. Finally, eye-gaze is a critical source of non-verbal information about human task and motion intent in human-robot collaboration, traffic maneuver prediction, spatial cognition or sign placement \cite{doshi2009roles,palinko2016robot,admoni2017social,kiefer2019eye,chadalavada2020bi}.

Existing datasets of human trajectories, commonly used in the literature \cite{rudenko2019human}, %include ETH \cite{pellegrini2009you}, UCY \cite{lerner2007crowds}, VIRAT \cite{oh2011large}, KITTI \cite{Geiger2012CVPR}, Edinburgh \cite{majecka2009statistical}, ATC \cite{brscic2013person}, Stanford Drone Dataset \cite{robicquet2016learning},
%Daimler Tracks Dataset \cite{schneider2013gcpr},
%L-CAS \cite{yan2017online}, Central Station \cite{zhou2012understanding} and the KTH tracks dataset \cite{dondrup2015tracker}.
are summarized in Table \ref{tab:datasets}.
With the exception of \cite{brscic2013person,yan2017online,zhou2012understanding,dondrup2015tracker}, all datasets have been collected outdoors. Intuitively, patterns of human motion in indoor and outdoor environments are substantially different due to scope of the environment and typical intentions of people therein. Indoors people navigate in loosely constrained but cluttered spaces with multiple goal points and many ways (e.g. from different homotopy classes) to reach a goal. This is different from their behavior outdoors in either large obstacle-free pedestrian areas or relatively narrow sidewalks, surrounded by various kinds of walkable and non-walkable surfaces. Among the indoor recordings, only \cite{yan2017online,dondrup2015tracker} introduce a robot, navigating in the environment alongside humans. However, recording only from on-board sensors limits visibility and consequently restricts the perception radius. Furthermore, ground truth positions of the recorded agents in all prior datasets were estimated from RGB(-D) or laser data. On the contrary, we directly record the position of each person using a motion capture system, thus achieving higher accuracy of the ground truth data and complete coverage of the working environment at all times. Moreover, our dataset contains many additional contextual cues, such as social roles and groups of people, head orientations and gaze directions.

%In particular, we aimed to address the the limitations of the existing motion trajectories datasets (see Sec.~\ref{sec:related_work:datasets}), posing the following requirements on the experiment:
%\begin{itemize}
%	\item Indoor experiment
%	\item Obstacles in the environment
%	\item Robot moving alongside with humans
%	\item Long trajectories, continuously tracked for 10+ seconds
%	\item Goal-oriented motion betweek several locations in the environment
%	\item Non-homogeneous agents, e.g. different motion patterns conditioned on the social role
%	\item Recording head orientation eye tracking data 
%\end{itemize}

\section{Data Collection Procedure}
\label{sec:experiment}

In order to collect motion data relevant for a broad spectrum of research areas, we have designed a controlled scenario that encourages social interactions between individuals, groups of people and with the robot. The interactive setup assigns social roles and tasks so as to imitate typical activities found in populated spaces such as offices, train stations, shopping malls or airports. Its goal is to motivate participants to engage into natural and purposeful motion behaviors as well as to create a rich variety of unscripted interactions. In this section we detail the system setup and the data collection procedure.

\subsection{System Setup}

\begin{figure}[t]
	\begin{center}
		\vspace{5pt}
		\includegraphics[width=0.999\columnwidth]{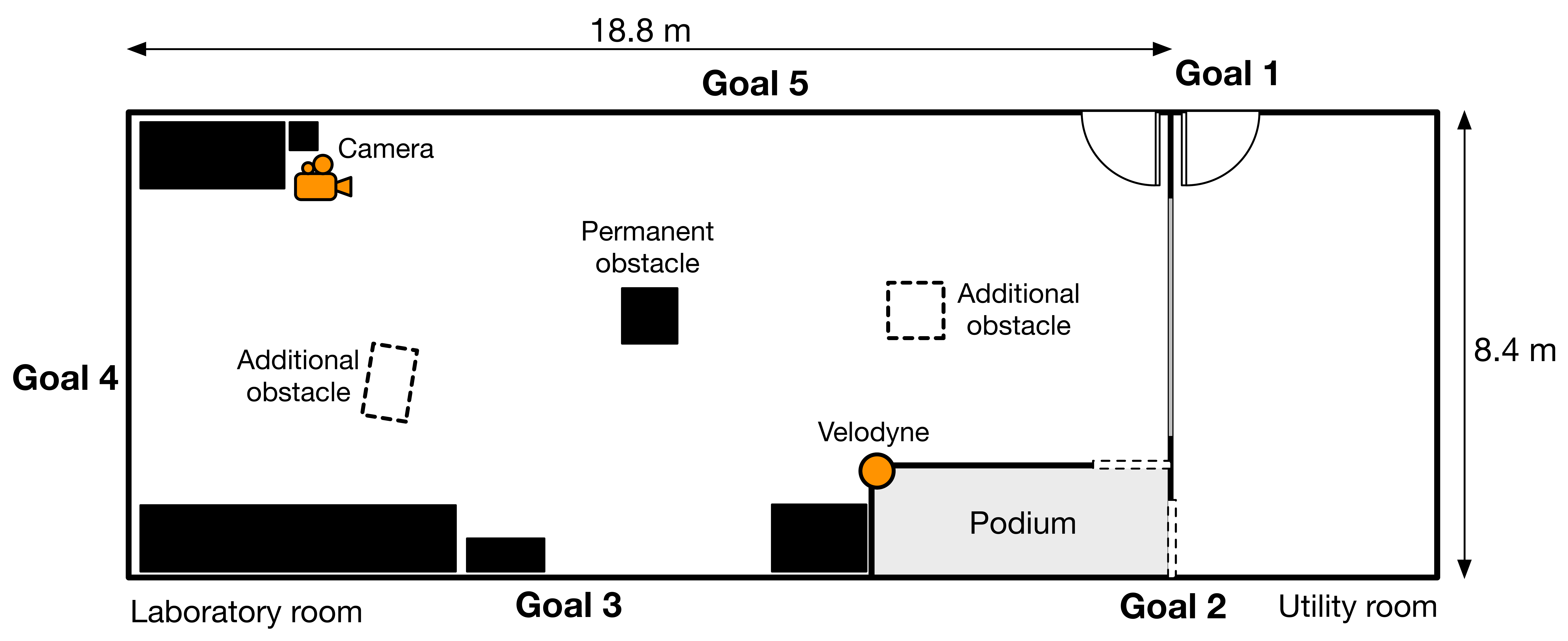}
	\end{center}
	\vspace{-8pt}
	\caption{Overview of the environment. The Qualisys motion tracking system is installed in a laboratory room, which is mostly empty except for some shelves and equipment along the walls. A permanent obstacle in the middle of the room is present in all recordings, while additional obstacles are only placed in the ``Three obstacles'' scenario (see Sec.~\ref{sec:experiment_description} for details). The position of the camera is shown in the top left corner, and the position of the Velodyne in the bottom right.}
	\label{fig:laboratory}
	%\vspace{-12pt}
\end{figure}

Data collection was performed in a spacious laboratory room of $8.4\times$\SI{18.8}{\metre} and the adjacent utility room, separated by a glass wall (see the overview in Fig.~\ref{fig:laboratory}). The laboratory room, where the motion capture system is installed, is mostly empty to allow for maneuvering of large groups, but also includes several constrained areas where obstacle avoidance and the choice of homotopy class is necessary. Goal positions are placed to force navigation along the room and generate frequent interactions in its center, while the placement of obstacles prevents walking between goals on a straight line.

\begin{figure}[t]
	\begin{center}
		\vspace{5pt}
		\includegraphics[width=0.49\columnwidth]{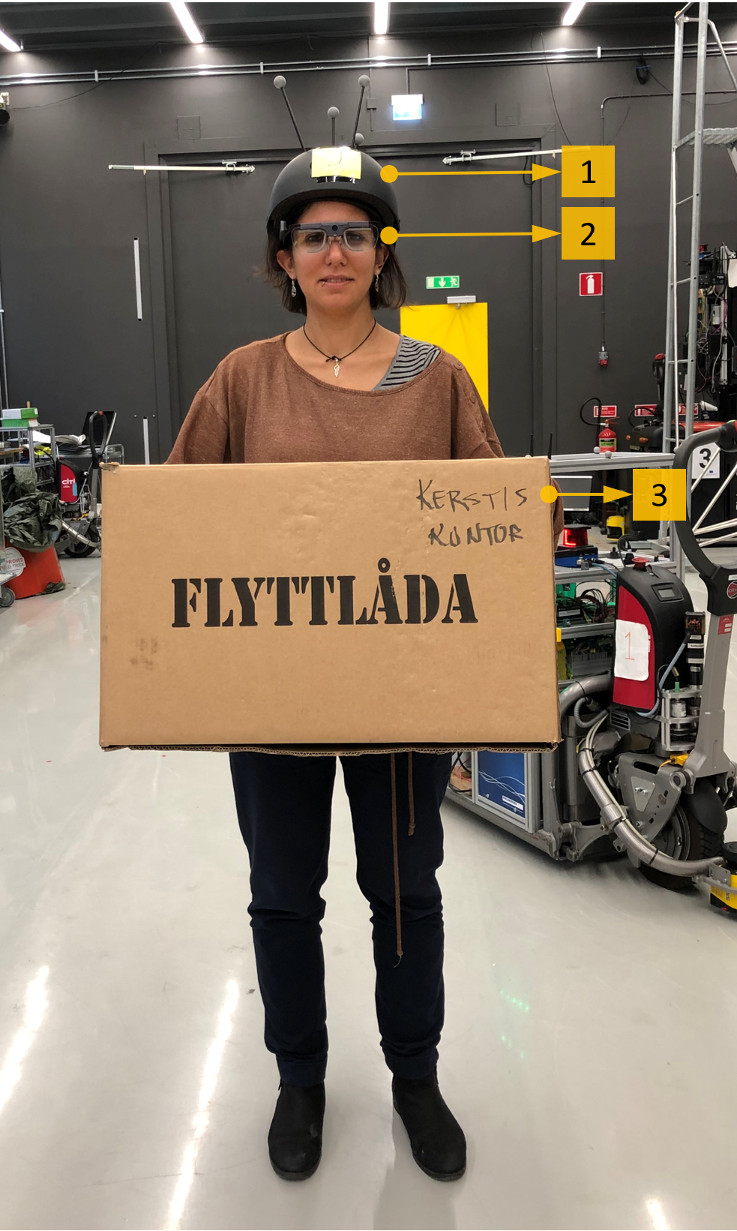}
		\includegraphics[width=0.33\columnwidth]{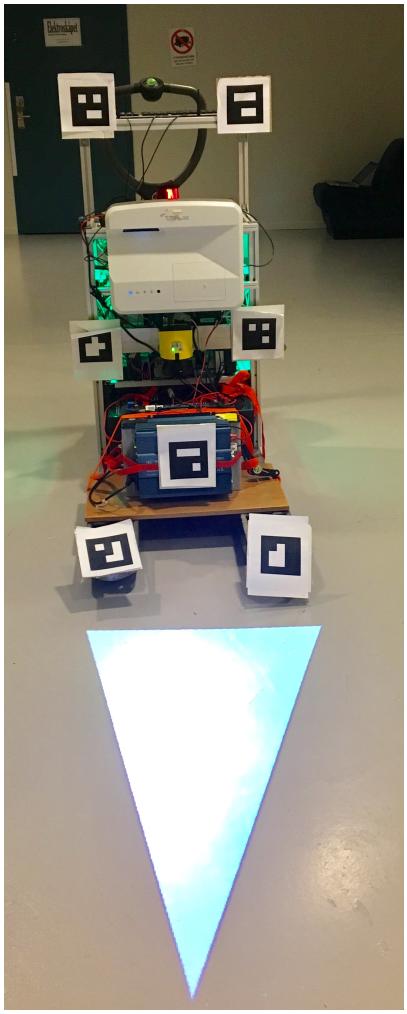}
	\end{center}
	\vspace{-8pt}
	%\caption{Equipment used in the experiment: {\bf Top left:} Bicycle helmet with mocap tracking markers. {\bf Top right:} Linde CitiTruck robot. {\bf Bottom left:} eye-traking glasses. {\bf Bottom right:} something else we want to show from the experiment. }
	\caption{Equipment used in our data collection: {\bf Left:} (1) bicycle helmet with mocap tracking markers, (2) Tobii Pro Glasses, (3) boxes which were carried by the participants as a part of the tasks. {\bf Right:} Linde CitiTruck robot projecting its current motion intent on the floor.}
	\label{fig:equipement}
	%\vspace{-12pt}
\end{figure}

To track the motion of the agents we used the Qualisys Oqus 7+ motion capture system\footnote{\url{https://www.qualisys.com/hardware/5-6-7/}} with 10 infrared cameras, mounted on the perimeter of the room. The motion capture system covers the entire room volume apart from the most right part close to the podium entrance -- a negligible loss due to the focus on the central part of the room. The system tracks small reflective markers at \SI{100}{\Hz} with spatial discretization of \SI{1}{\milli\metre}. The coordinate frame origin is on the ground level in the middle of the room. For people tracking, the markers have been arranged in distinctive 3D patterns on the bicycle helmets, shown in Fig.~\ref{fig:equipement}. The motion capture system was calibrated beforehand with an average residual tracking error of \SI{2}{\milli\metre}, and each helmet, as well as the robot, was defined in the system as a unique rigid body of markers, yielding its 6D head position and orientation. Each participant was assigned an individual helmet for all recording sessions, labeled 2 to 10. Helmet 1 was not used in this data collection.

%To track the gaze direction we used 4 eye-tracking glasses, worn by the same 4 participants for the duration of the experiment. 
%modified by Ravi
For acquiring eye gaze data we used four mobile eye-tracking headsets
%: three Pupil labs eye-tracking headsets\footnote{\url{https://pupil-labs.com/products/core/}} and one Tobii Pro Glasses\footnote{\url{https://www.tobiipro.com/product-listing/tobii-pro-glasses-2/}}, 
worn by four participants (helmet numbers 3, 6, 7, and 9 respectively). However, in this dataset we only include data from one headset (Tobii Pro Glasses), worn by the participant with helmet 9. 
%Due to technical failures pertaining to eye-tracking headsets, data from three Pupil labs eye-tracking headsets was not successfully recorded and hence excluded from this dataset.
The gaze sampling frequency of Tobii Pro Glasses is \SI{50}{\Hz}.  It also has a scene camera which records the video at 25 fps. A gaze overlaid version of this video is included in this dataset. We synchronized the clocks of each machine (the Qualisys system, the stationary Velodyne sensor and the eye-tracking glasses) with the same NTP time server. Finally, we recorded a video of the environment from a stationary camera, mounted in a corner of the room.

The robot, used in our data collection, is a small forklift Linde CitiTruck robot with a footprint of $1.56\times$\SI{0.55}{\metre} and \SI{1.17}{\metre} high, shown in Fig.~\ref{fig:equipement}. It was programmed to move in a socially unaware manner, following a pre-defined path around the room and adjusting neither its speed nor trajectory to account for surrounding people. For safety reasons, the robot was navigating with a maximal speed of \SI{0.34}{\metre\per\second} and projecting its current motion intent on the floor in front of it using a mounted beamer \cite{chadalavada2020bi}. A dedicated operator was constantly monitoring the environment from a remote workstation to stop the robot in case of an emergency. The participants were made aware of the emergency stop button on the robot should they be required to use it. 

%\subsection{Participant Instructions/Priming/...}

\subsection{Scenario Description and Participants' Priming}
\label{sec:experiment_description}

\begin{figure}[t]
	\begin{center}
		\vspace{5pt}
		\includegraphics[width=0.999\columnwidth]{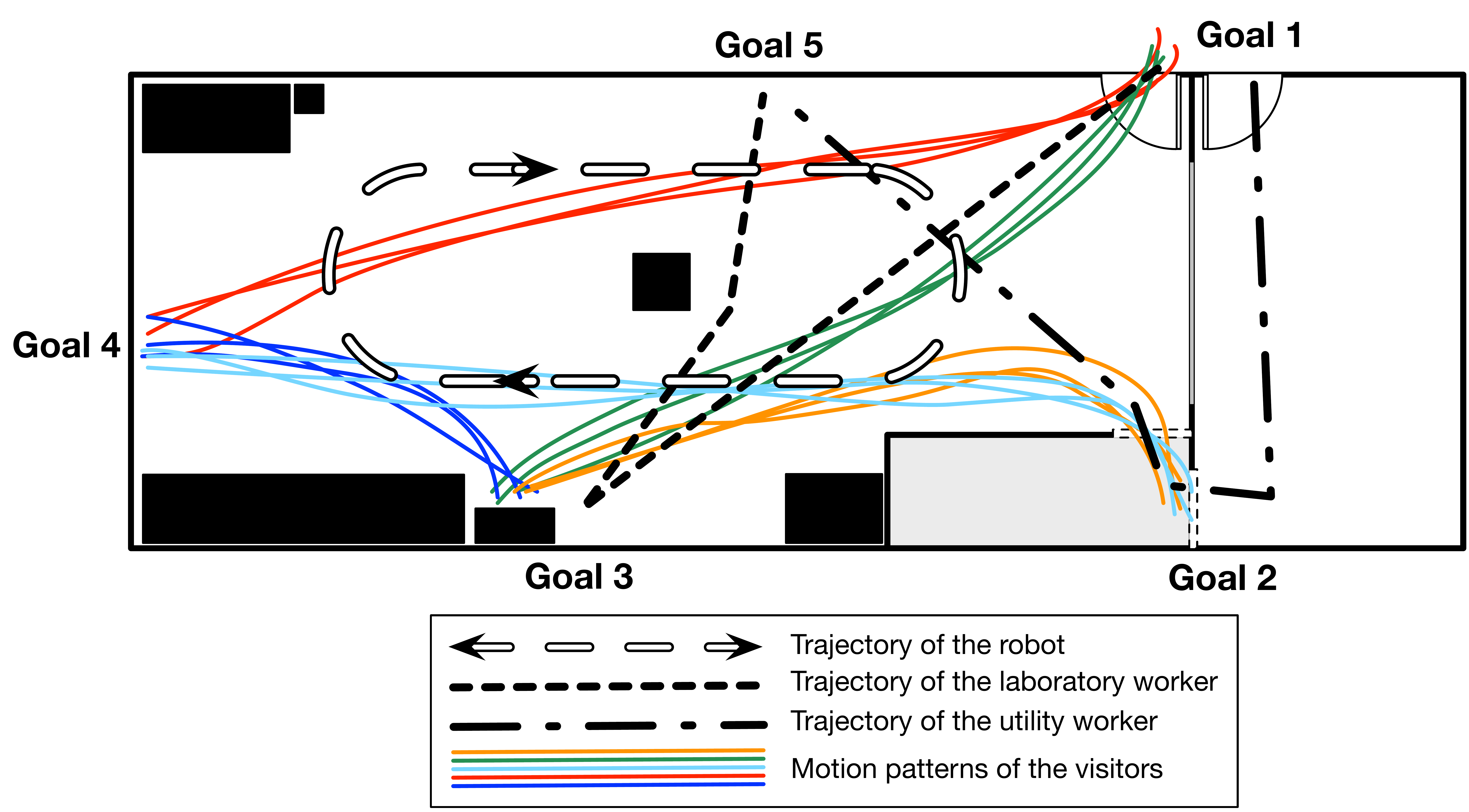}
	\end{center}
	\vspace{-8pt}
	\caption{Roles of the participants and their expected motion patterns. \emph{Visitors}, walking alone and in groups, are instructed to navigate between goals 1,2,3 and 4. Their motion patterns are shown with colored solid lines. The \emph{laboratory worker}, whose waiting position is at goal 3, picks up an incoming box at goal 1, registers its ID at goal 3 and then places it at goal 5. The \emph{utility worker}, whose waiting position is at goal 2, picks up the box at goal 5, registers it at goal 2 with a new ID and places it at goal 1. The patterns of both workers are shown with dotted lines. The trajectory of the \emph{robot}, circulating around the obstacle in the middle of the room, is shown with a thick hollow line.}
	\label{fig:experiment-trajectories}
	%\vspace{-12pt}
\end{figure}

During the data collection the participants performed simple tasks, which required walking between several goal positions. To increase the variety of motion, interactions and behavioral patterns, we introduced several roles for the participants and created individual tasks for each role, summarized in Fig.~\ref{fig:experiment-trajectories}.

The first role is a \emph{visitor}, navigating alone and in groups of up to 5 people between four goal positions in the room. At each goal they take a random card, indicating the next target. As each group was instructed to travel together, they only take one card at a time. We asked the visitors to talk and interact with the members of their group during the data collection, and changed the structure of groups every 4-5 minutes. There are 6 visitors in our recording. The second role is a \emph{worker}, whose task is to receive and carry large boxes between the laboratory and the utility room. The workers wear a yellow reflective vest. There are 2 workers in our recording, one carrying the boxes from the laboratory to the unity room, and the other vice versa. The third role is the \emph{inspector}. An inspector is navigating alone between many additional targets in the environment, indicated by a QR-code, in no particular order and stops at each target to scan the code. We have one inspector in our recording.

There are several points to motivate the introduction of the social roles. Firstly, with the motion of the visitors and the workers we introduce distinctive motion patterns in the environment, while the cards and the tasks make the motion focused, goal-oriented and prevent random wandering. However, the workers' tasks allocation is such that at some points idle standing/wandering behavior is also observed, embedded in their cyclical activity patterns. Furthermore, we expect that the visitors navigating alone, in groups and the workers who carry heavy boxes exhibit distinctive behavior, therefore the grouping information and the social role cue (reflective vest) may improve the intention and trajectory prediction. Finally, motion of the inspector introduces irregular patterns in the environment, distinct from the majority of the visitors.

We prepared three scenarios for data collection with different numbers of obstacles and motion state of the robot. In the first scenario, the robot is placed by a wall and not moving, and the environment has only one obstacle (see the layout in Fig.~\ref{fig:laboratory}). The second scenario introduces the moving robot, navigating around the obstacle (the trajectory of the robot is depicted in Fig.~\ref{fig:experiment-trajectories}). The third scenario features an additional obstacle and a stationary robot in the environment (see Fig.~\ref{fig:laboratory} with additional obstacles). We denote these recording scenarios as \emph{One obstacle}, \emph{Moving robot} and \emph{Three obstacles}, accordingly. In each scenario the group structure for the visitors was reassigned 4-5 times. Between the scenarios, the roles were also reassigned. A summary of the scenarios and durations is given in Table~\ref{tab:experiments}.

\begin{table}[t!]
	\center
	\vspace{5pt}
	\footnotesize{
		\begin{tabular}{ | p{1.42cm}| p{1.89cm}| p{1.22cm}| p{0.98cm}| p{0.95cm}|}
			\hline  
			\textbf{Scenario, round} & \textbf{Visitors, groups} Helmet ID 2--10 & \textbf{Workers} Utility, \hfill lab & \textbf{Inspector} & \textbf{Duration}\\ \hline
			One \hfill 1 & 6,7,5 \hspace{1pt} + \hspace{1pt} 8,2,4 & 3 \hfill 9  & 10 & 368 sec \\
			obstacle \hfill 2 & 2,5,6,7 \hspace{1pt} + \hspace{1pt} 8,4 & 3 \hfill 9  & 10 & 257 sec \\
			\hfill 3 & 6,7,8  +  4,5  +  2 & 3 \hfill 9  & 10 & 275 sec \\
			\hfill 4 & 2,4,5,7,8 \hspace{1pt} + \hspace{1pt} 6 & 3 \hfill 9  & 10 & 315 sec \\ \hline
			Moving \hfill 1 & 4,5,6 \hspace{1pt} + \hspace{1pt} 3,7,9 & 2 \hfill 8  & 10 & 281 sec \\
			robot \hfill 2 & 3,5,6,9 \hspace{1pt} + \hspace{1pt} 7,4 & 2 \hfill 8  & 10 & 259 sec \\
			\hfill 3 & 5,7,9  +  4,6  +  3 & 2 \hfill 8  & 10 & 286 sec \\
			\hfill 4 & 3,5,6,7,9 \hspace{1pt} + \hspace{1pt} 4 & 2 \hfill 8  & 10 & 279 sec \\
			\hfill 5 & 3,6 + 4,9 + 5,7  & 2 \hfill 8  & 10 & 496 sec \\ \hline
			Three \hfill 1 & 2,3,8 \hspace{1pt} + \hspace{1pt} 6,7,9 & 5 \hfill 4  & 10 & 315 sec \\
			obstacles \hfill 2 & 2,8,9 \hspace{1pt} + \hspace{1pt} 3,6,7 & 5 \hfill 4  & 10 & 290 sec \\
			\hfill 3 & 2,3,7  +  8,9  +  6 & 5 \hfill 4  & 10 & 279 sec \\
			\hfill 4 & 2,3,6,7,9 \hspace{1pt} + \hspace{1pt} 8 & 5 \hfill 4  & 10 & 277 sec \\ \hline
		\end{tabular}
	}
	\vspace{10pt}
	\caption{Role assignment and recording duration in the three scenarios of our data collection: (i)~One~obstacle, (ii)~Moving~robot, (iii)~Three~obstacles}
	\label{tab:experiments}
	\vspace{-8pt}
\end{table}

Each round of data collection started with the participants, upon command, beginning to execute their tasks. The round lasted for approximately four minutes and ended with another call from the moderator. To avoid artificial and unnatural motion due to knowing the true purpose of the data collection, we told the participants that our goal is to validate the robot's perceptive abilities, while the motion capture data will be used to compare the perceived and actual positions of humans. Participants were asked not to communicate with us during the recording. For safety and ethical reasons, we have instructed participants to act carefully near the robot, described as ``autonomous industrial equipment'' which does not stop if someone is in its way. 
%The experimental procedure was approved by the university ethics board.
An ethics approval was not required for our data collection as per institutional guidelines and the Swedish Ethical Review Act (SFS number: 2003:460). Written informed consent was obtained from all participants. Due to the relatively low weight of the robot and the safety precautions taken, there was no risk of harm to participants.
%Prior to the experiment, participants were offered a complimentary lunch and signed the consent form. After the experiments participants filled out a questionnaire. (details ... ...). The entire experiment took approximately 2 hours.

%\subsection{Implementation details}

%There were 9 participants, whom we divided into 2 workers, 6 visitors and one person moving between numerous additional goals in the environment.

\section{Results and Analysis}
\label{sec:results}

\begin{figure}[t]
	\begin{center}
		\vspace{5pt}
		\includegraphics[width=0.98\columnwidth]{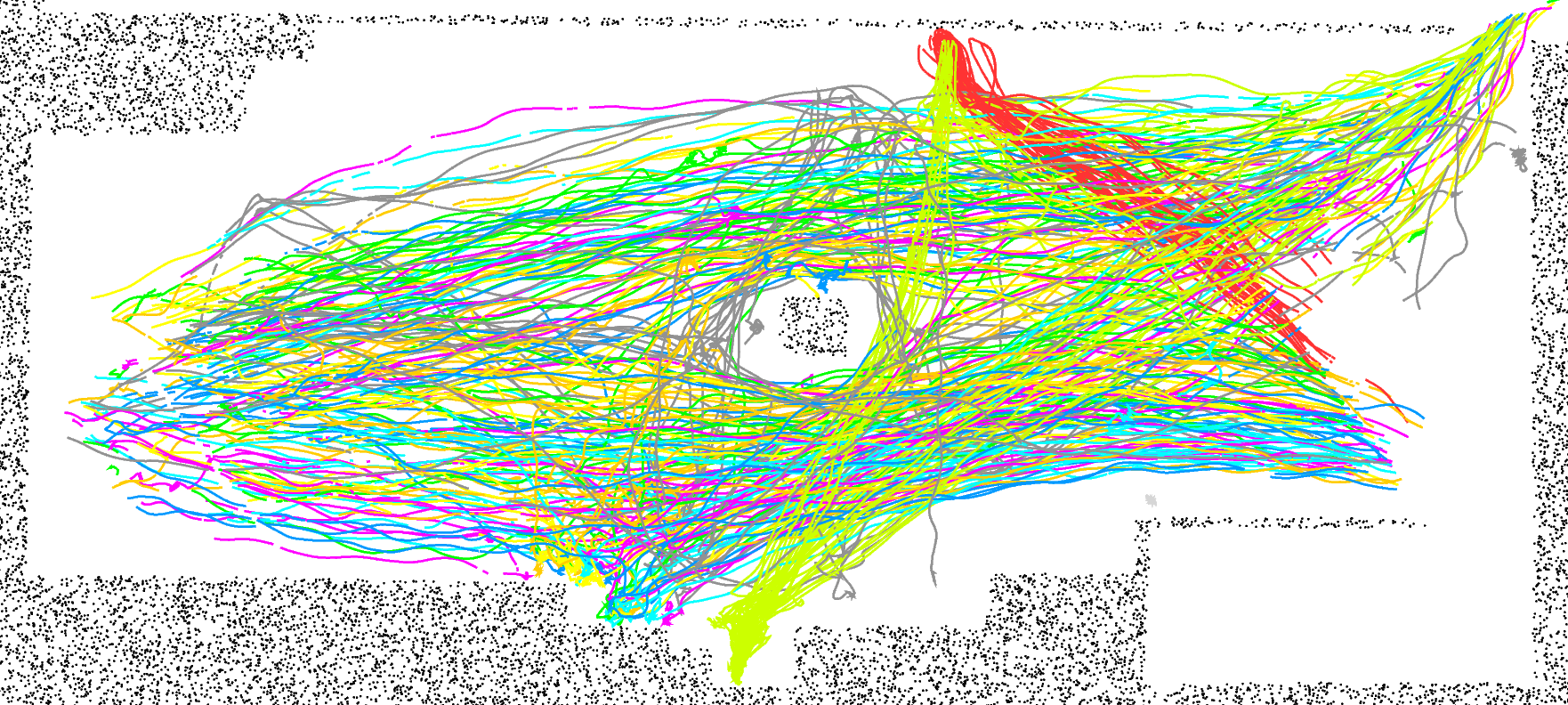}
		\includegraphics[width=0.98\columnwidth]{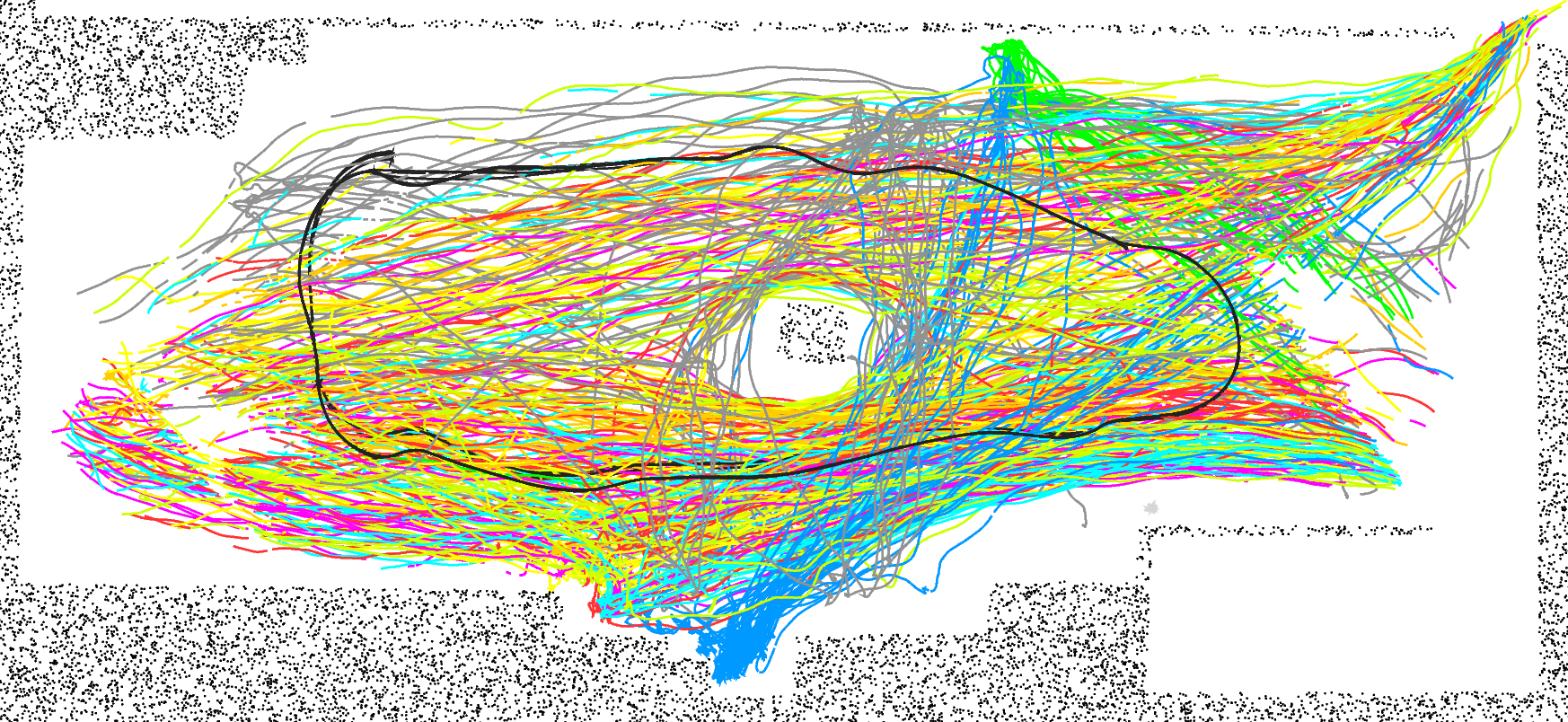}
		\includegraphics[width=0.98\columnwidth]{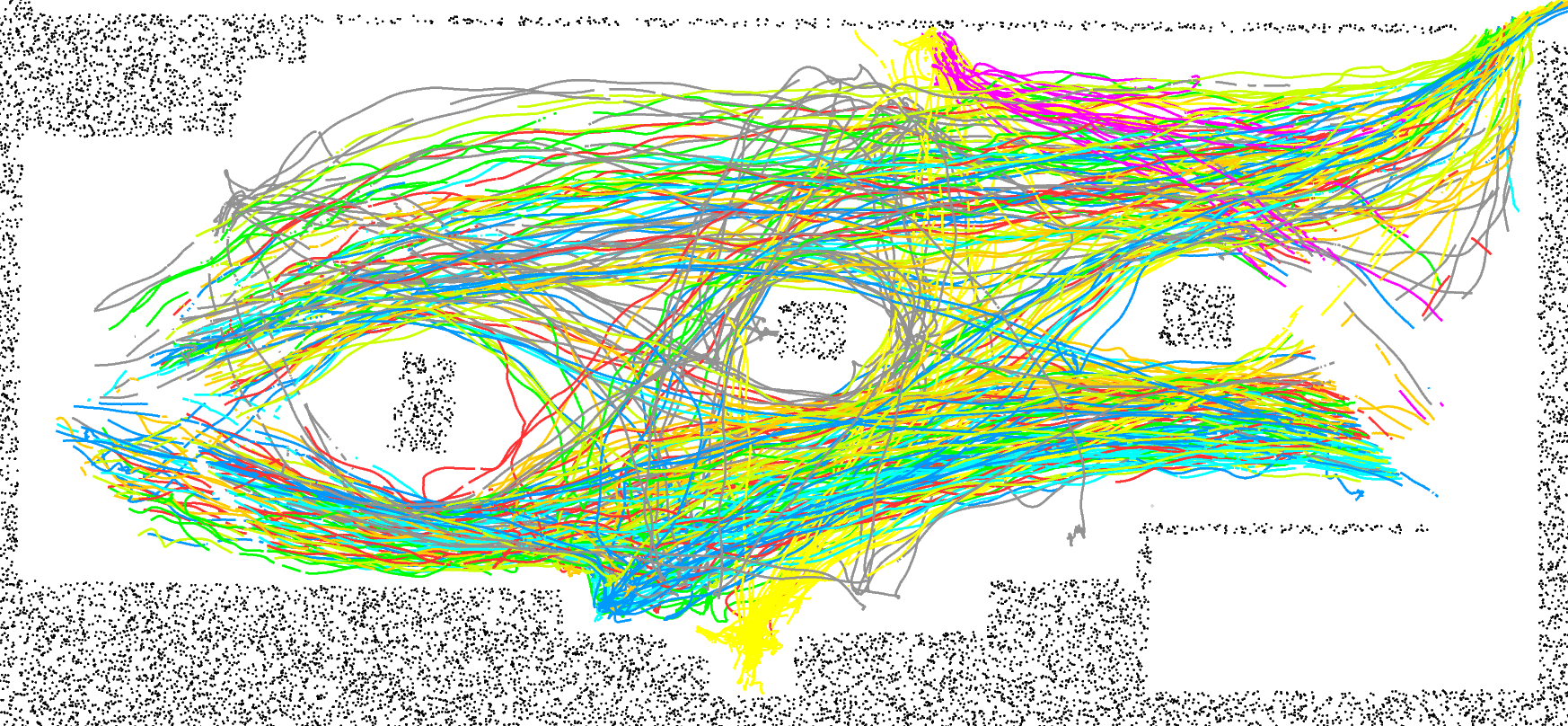}
	\end{center}
	\vspace{-8pt}
	\caption{Trajectories of the participants and the robot, recorded in the ``One obstacle'' scenario {\bf (top)}, ``Moving robot'' scenario {\bf (middle)} and ``Three obstacles'' scenario {\bf (bottom)}. The robot's path in the middle image is shown in {\bf black}.}
	\label{fig:recorded_trajectories}
	%\vspace{-12pt}
\end{figure}

\subsection{Data Description}

The TH\"OR dataset includes over 60 minutes of motion in 13 rounds of the three scenarios. The recorded data in \texttt{.mat}, \texttt{.bag} and \texttt{.tsv} format contains over 395k frames at \SI{100}{\Hz}, 2531k human detections and 600+ individual and group trajectories between the goal positions. For each detected person the 6D position and orientation of the helmet in the global coordinate frame is provided. Furthermore, the dataset includes the map of the static obstacles, goal coordinates and grouping information. We also share the Matlab scripts for loading, plotting and animating the data. Additionally, the eye gaze data is available for one of the participants (Helmet 9), as well as Velodyne scans from a static sensor and the recording from the camera. We thoroughly inspected the motion capture data and manually cleaned it to remove occasional helmet ID switches and recover several lost tracks. Afterwards we applied an automated procedure to restore the lost positions of the helmets from incomplete set of recognized markers. In Fig.~\ref{fig:recorded_trajectories} we show the summary of the recorded trajectories. %Additionally, in Fig.~\ref{...} we show the distributions of orientations and velocities in the CLIFF-map of the environment \cite{kucner2017enabling}.

\subsection{Baselines and Metrics}

The TH\"OR dataset is recorded using a motion capture system, which yields more consistent tracking and precise estimation of the ground truth positions and therefore higher quality of the trajectories, compared to the human detections from RGB-D or laser data, typically used in existing datasets. For the quantitative analysis of the dataset, we compare the recorded trajectories to the several datasets which are often used for training and evaluation of motion predictors for human environments \cite{rudenko2019human}. The popular ETH dataset \cite{pellegrini2009you} is recorded outdoors in a pedestrian zone with a stationary camera facing downwards and manually annotated at \SI{2.5}{\Hz}. The Hotel sequence, used in our comparison, includes the coordinates of the 4 common goals in the environment and group information for walking pedestrians. The ATC dataset \cite{brscic2013person} is recorded in a large shopping mall using multiple 3D range sensors at $\sim$\SI{26}{\Hz} over an area of \SI{900}{\metre}$^2$. This allows for long tracking durations and potential to capture interesting interactions between people. In addition to positions it also includes facing angles. In this comparison we used the recordings from 24th and 28th of October and 14th of November. The Edinburgh dataset \cite{majecka2009statistical} is recorded in a university campus yard using a camera facing down with variable detection frequency, on average \SI{9}{\Hz}. For comparison we used the recordings from 27th of September, 16th of December, 14th of January and 22nd of June.

For evaluating the quality of recorded trajectories we propose several metrics:

\begin{enumerate}
	\item \emph{Tracking duration} (\si{\second}): average length of continuous observations of a person, higher is better.
	\item \emph{Trajectory curvature} (\si{\per\metre}): global curvature of the trajectory $\mathcal{T}$, caused by maneuvering of the agents in presence of static and dynamic obstacles, measured on \SI{4}{\second} segments based on the first $\mathcal{T}_t=(x_1,y_1)$, middle $\mathcal{T}_{t+2 s}=(x_2,y_2$) and last $\mathcal{T}_{t+4 s}=(x_3,y_3)$ points of the interval: $K(\mathcal{T}_{t:t+4 s}) = |\frac{2(x_2-x_1)(y_3-y_1)-(x_3-x_1)(y_2-y_1)}{ ||x_2-x_1,y_2-y_1||~||x_3-x_1,y_3-y_1||~||x_3-x_2,y_3-y_2||}|$. The choice of \SI{4}{\second} path segments is motivated by the standard motion prediction horizon in the related work \cite{alahi2016social}. Higher curvature values correspond to more challenging, non-linear paths.
	\item \emph{Perception noise} (\si{\metre\per\s\squared}): under the assumption that people move on smooth, not jerky paths, we evaluate local distortions of the recorded trajectory $\{\mathcal{T}_t\}_{t=1\dots M}$ of length $M$, caused by the perception noise of the mocap system as the average absolute acceleration: $\frac{1}{M}\sum_{t=1}^{M} |\ddot{\mathcal{T}_t}|$. Less noise is better.
	\item \emph{Motion speed} (\si{\metre\per\second}): mean and standard deviation of velocities in the dataset, measured on \SI{1}{\second} intervals. If the effect of perception noise on speed is negligible, higher standard deviation means more diversity in behavior of the observed agents, both in terms of individually preferred velocity and compliance with other dynamic agents.
	\item \emph{Minimal distance between people} (\si{\metre}): average minimal euclidean distance between two closest observed people. This metric indicates the density of the recorded scenarios, lower values correspond to more crowded environments.
\end{enumerate}

\begin{table}[t!]
	\center
	\vspace{5pt}
	\footnotesize{
		\begin{tabular}{ | p{1.26cm}| p{1.30cm}| p{1.30cm}| p{1.30cm}| p{1.30cm}|}
			\hline  
			\textbf{Metric} & \textbf{TH\"OR} & \textbf{ETH} & \textbf{ATC} & \textbf{Edinburgh}\\ \hline
			Tracking duration [\si{\second}] & $16.7 \pm 14.9$ & $9.4  \pm  5.4$ & $39.7  \pm 64.7$ & $10.1 \pm  12.7$ \\ \hline
			Trajectory curvature [\si{\per\metre}] & $1.9 \pm 8.8$ & $0.18 \pm 1.48$ & $0.84 \pm 1.43$ & $1 \pm 3.9$ \\ \hline
			Perception noise [\si{\metre\per\s\squared}] & $0.12$ & $0.19$ & $0.48$ & $0.81$ \\ \hline
			Motion speed [\si{\metre\per\second}] & $0.81 \pm 0.49$ & $1.38 \pm 0.46$ & $1.04 \pm 0.46$ & $1.0 \pm 0.64$ \\ \hline	
			Min.~dist. between people~[\si{\metre}] & $1.54 \pm 1.60$ & $1.33 \pm 1.39$ & $0.61 \pm 0.16$ & $3.97 \pm 3.5$ \\ \hline
			%Sampling frequency & 100 Hz & 2.5 Hz & 26 Hz & 9 Hz \\ \hline
		\end{tabular}
	}
	\vspace{10pt}
	\caption{Comparison of the datasets}
	\label{tab:results}
	\vspace{-8pt}
\end{table}

\begin{figure*}[t]
	\begin{center}
		\vspace{5pt}
		\includegraphics[width=1.99\columnwidth]{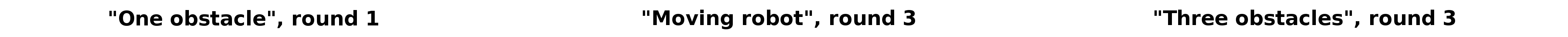} 
		\includegraphics[width=0.666666\columnwidth]{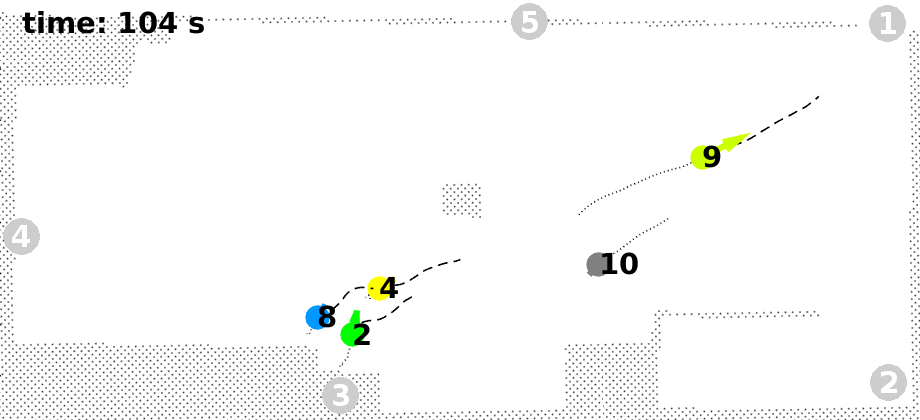} 
		\includegraphics[width=0.666666\columnwidth]{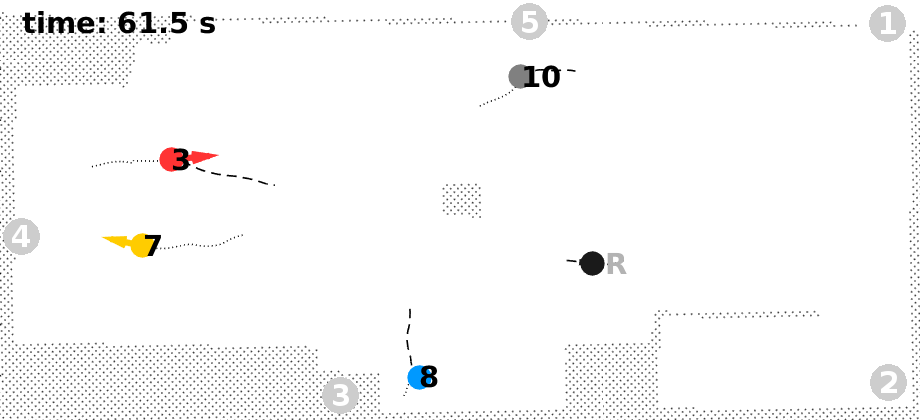}
		\includegraphics[width=0.666666\columnwidth]{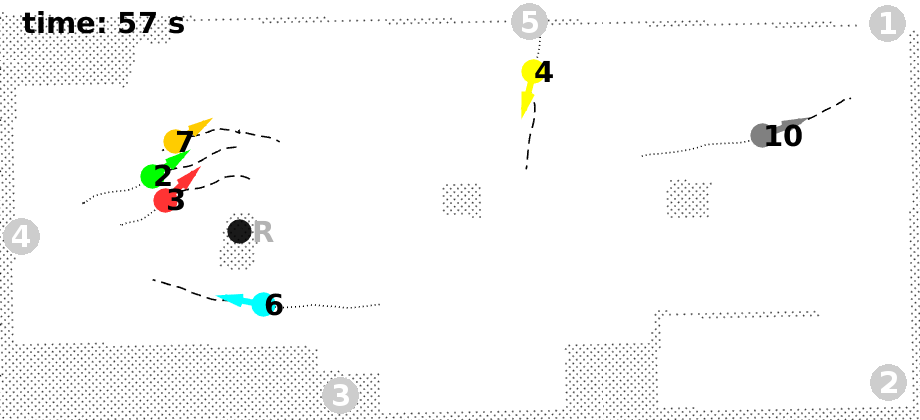} \\
		\includegraphics[width=0.666666\columnwidth]{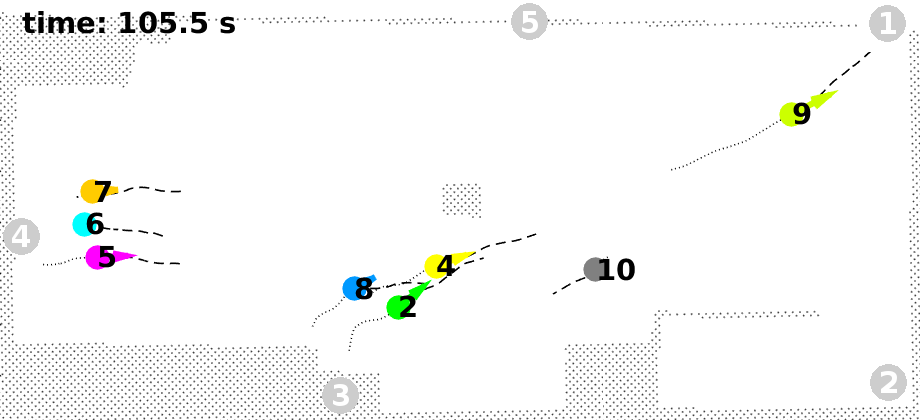} 
		\includegraphics[width=0.666666\columnwidth]{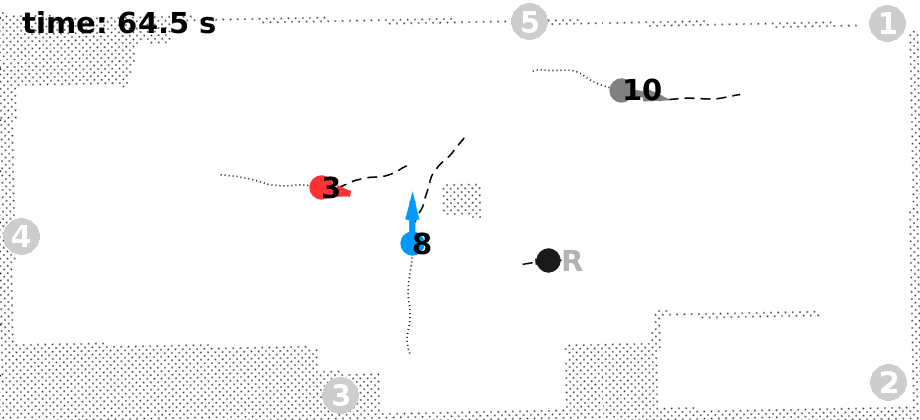}
		\includegraphics[width=0.666666\columnwidth]{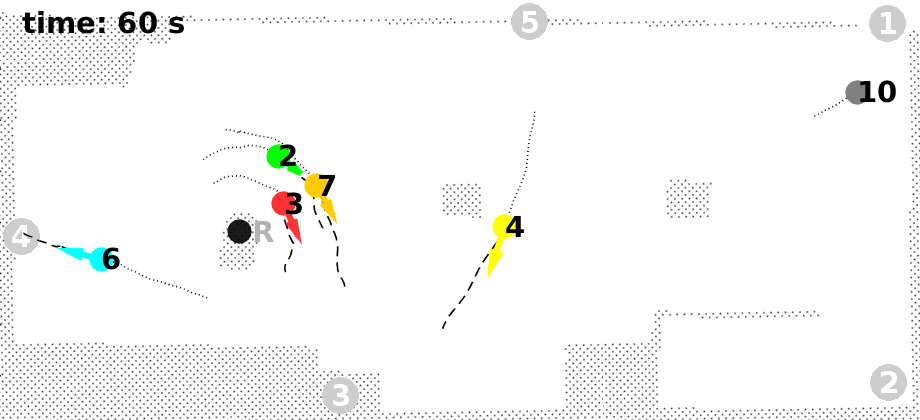} \\
		\includegraphics[width=0.666666\columnwidth]{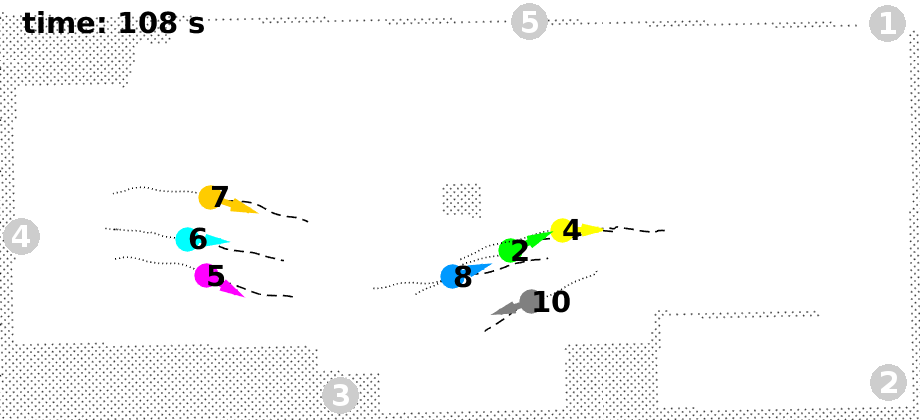} 
		\includegraphics[width=0.666666\columnwidth]{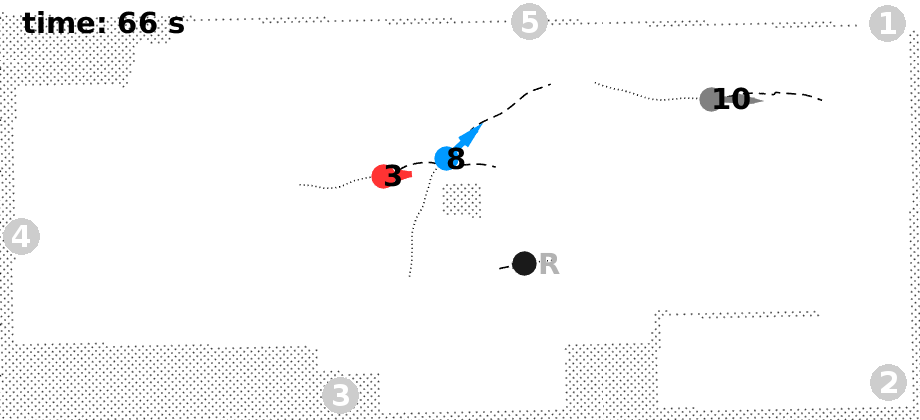}
		\includegraphics[width=0.666666\columnwidth]{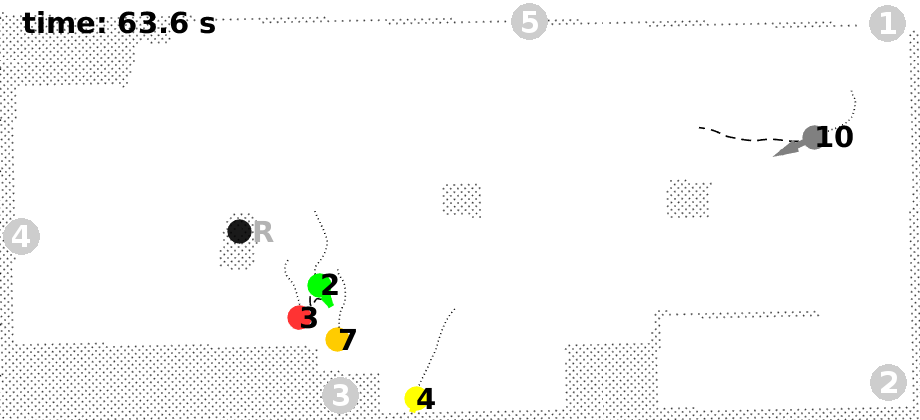} \\
		\includegraphics[width=0.666666\columnwidth]{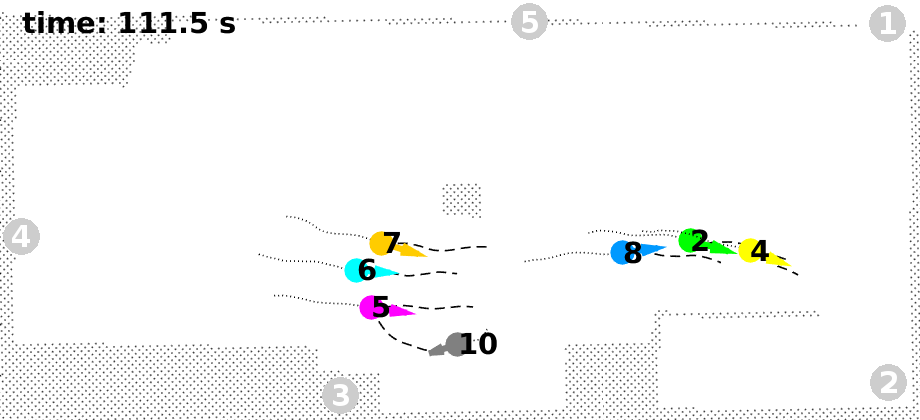} 
		\includegraphics[width=0.666666\columnwidth]{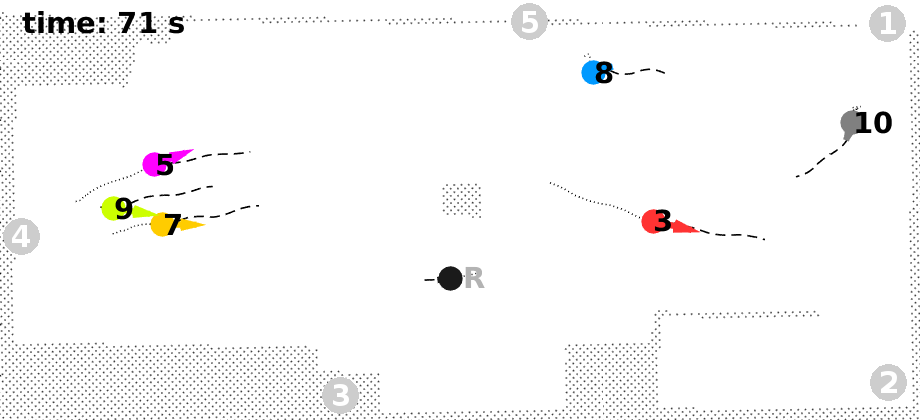}
		\includegraphics[width=0.666666\columnwidth]{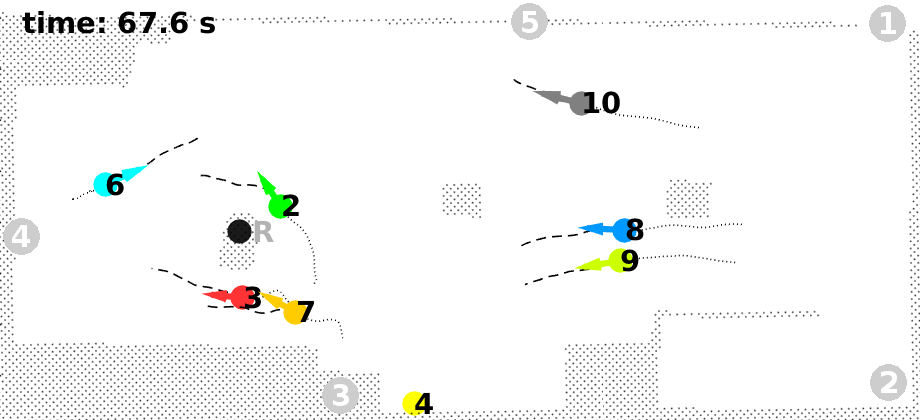} \\
		\includegraphics[width=0.666666\columnwidth]{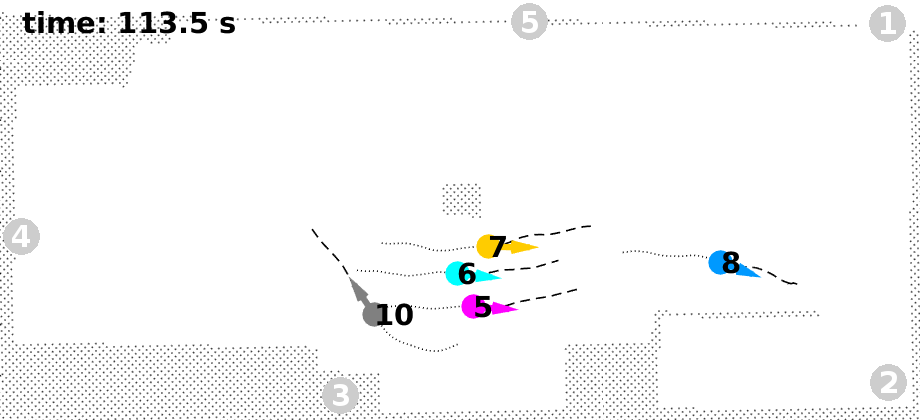}
		\includegraphics[width=0.666666\columnwidth]{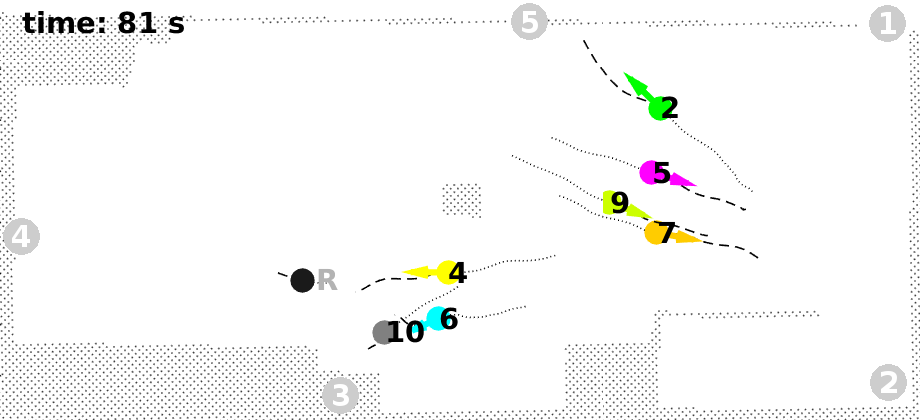}
		\includegraphics[width=0.666666\columnwidth]{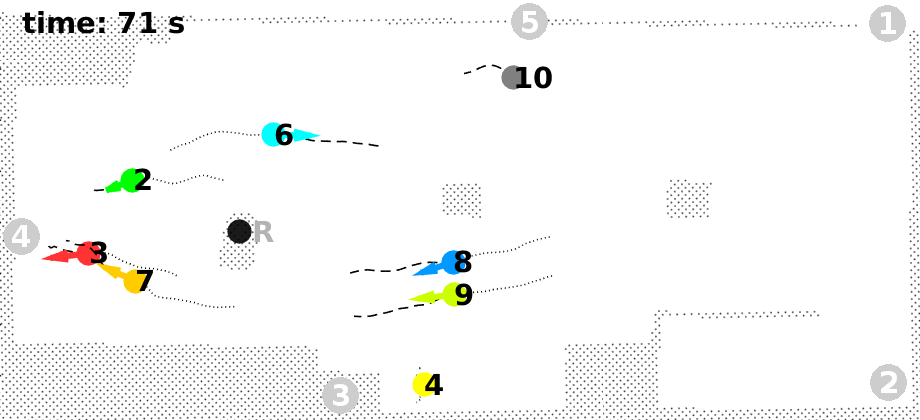} \\
	\end{center}
	\vspace{-8pt}
	\caption{Social interactions in the TH\"OR dataset with color-coded positions of the observed people. The current velocity is shown with an arrow of corresponding length and direction. The past and the future 2 s trajectories are shown with dotted and dashed lines respectively. Goal locations are marked with gray circles. {\bf Left column:} at 104 sec the group (2,4,8) starts moving from the goal point, taking the \emph{line formation} in the constrained space due to the presence of standing person 10. Later, at 111.5 sec, person 10 has to adjust the path and slow down while the group (5,6,7) proceeds in the \emph{V formation} \cite{moussaid2010walking}, engaged in communication. {\bf Middle column:} person 8 is leaving the resting position at 61.5 sec and adapts the path to account for the motion of the robot, taking a detour from the optimal way to reach the goal 5. At 66 seconds person 8 crosses person 3, who has to slow down, as compared to the velocity at time 61.5 and 71. The same maneuver of taking a detour due to the presence of the robot is performed by the group (5,7,9) at time 71. {\bf Right column:} Group (2,3,7), navigating in a constrained environment, at 57 sec has to make a detour around the obstacle while heading to goal 3. On the way back to goal 4 the group splits at 67.6 sec, and reunites later on.}
	\label{fig:samples}
	%\vspace{-12pt}
\end{figure*}
%{\bf Middle column:} person 9 departs from the goal position, adjusting the path to account for the group (5,6,7), moving in the opposite direction. Person 10 halts for 3 seconds and then heads to the exit. 

\subsection{Results}

The results of the evaluation are presented in Table~\ref{tab:results}. Our dataset has sufficiently long trajectories (on average \SI{16.7}{\second} tracking duration) with high curvature values ($1.9\pm 8.8$~\si{\per\metre}), indicating that it includes more human-human and human-environment interactions than the existing datasets. Furthermore, despite the much higher recording frequency, e.g. \SI{100}{\Hz} (TH\"OR) vs. $\sim$\SI{26}{\Hz} (ATC), the amount of perception noise in the trajectories is lower than in all baselines. The speed distribution of $\pm$\SI{0.49}{\metre\per\second} shows that the range of observed velocities corresponds to the baselines, while the lower average velocity in combination with a high average curvature confirms higher complexity of the recorded behaviors, because comfortable navigation in straight paths with constant velocity is not possible in presence of static and dynamic obstacles. Finally, the high variance of the minimal distance between people ($1.54\pm$\SI{1.60}{m} TH\"OR vs. $0.61\pm$\SI{0.16}{m} ATC) shows that our dataset features both dense and sparse scenarios, similarly to ETH and Edinburgh.

An important advantage of TH\"OR in comparison to the prior art is the availability of rich interactions between the participants and groups in presence of static obstacles and the moving robot. In this compact one hour recording we observe numerous interesting situations, such as accelerating to overtake another person; cutting in front of someone; halting to let a large group pass; queuing for the occupied goal position; group splitting and re-joining; choosing a sub-optimal motion trajectory from a different homotopy class due to a narrow passage being blocked; hindrance from walking towards each other in opposite directions. In Fig.~\ref{fig:samples} we illustrate several examples of such interactions.

\section{Conclusions}
\label{sec:conclusions}

In this paper we present a novel human motion trajectories dataset, recorded in a controlled indoor environment. Aiming at applications in training and benchmarking human-aware intelligent systems, we designed the dataset to include a rich variety of human motion behaviors, interactions between individuals, groups and a mobile robot in the environment with static obstacles and several motion targets. Our dataset includes accurate motion capture data at high frequency, head orientations, eye gaze directions, data from a stationary 3D lidar sensor and an RGB camera. Using a novel set of metrics for the dataset quality estimation, we show that it is less noisy and contains higher variety of behavior than the prior art datasets.

%\addtolength{\textheight}{-12cm}   % This command serves to balance the column lengths
                                  % on the last page of the document manually. It shortens
                                  % the textheight of the last page by a suitable amount.
                                  % This command does not take effect until the next page
                                  % so it should come on the page before the last. Make
                                  % sure that you do not shorten the textheight too much.

%%%%%%%%%%%%%%%%%%%%%%%%%%%%%%%%%%%%%%%%%%%%%%%%%%%%%%%%%%%%%%%%%%%%%%%%%%%%%%%%

%%%%%%%%%%%%%%%%%%%%%%%%%%%%%%%%%%%%%%%%%%%%%%%%%%%%%%%%%%%%%%%%%%%%%%%%%%%%%%%%

%%%%%%%%%%%%%%%%%%%%%%%%%%%%%%%%%%%%%%%%%%%%%%%%%%%%%%%%%%%%%%%%%%%%%%%%%%%%%%%%
%\section*{APPENDIX}

%Appendixes should appear before the acknowledgment.

\section*{Acknowledgments}

The authors would like to thank Martin Magnusson for his invaluable support with the motion capture system and Luigi Palmieri for insightful discussions.

%The preferred spelling of the word ÒacknowledgmentÓ in America is without an ÒeÓ after the ÒgÓ. Avoid the stilted expression, ÒOne of us (R. B. G.) thanks . . .Ó  Instead, try ÒR. B. G. thanksÓ. Put sponsor acknowledgments in the unnumbered footnote on the first page.

%%%%%%%%%%%%%%%%%%%%%%%%%%%%%%%%%%%%%%%%%%%%%%%%%%%%%%%%%%%%%%%%%%%%%%%%%%%%%%%%

\bibliographystyle{IEEEtran}
\footnotesize{
	\bibliography{bibliography}

% Generated by IEEEtran.bst, version: 1.14 (2015/08/26)
\begin{thebibliography}{10}
\providecommand{\url}[1]{#1}
\csname url@samestyle\endcsname
\providecommand{\newblock}{\relax}
\providecommand{\bibinfo}[2]{#2}
\providecommand{\BIBentrySTDinterwordspacing}{\spaceskip=0pt\relax}
\providecommand{\BIBentryALTinterwordstretchfactor}{4}
\providecommand{\BIBentryALTinterwordspacing}{\spaceskip=\fontdimen2\font plus
\BIBentryALTinterwordstretchfactor\fontdimen3\font minus
  \fontdimen4\font\relax}
\providecommand{\BIBforeignlanguage}[2]{{%
\expandafter\ifx\csname l@#1\endcsname\relax
\typeout{** WARNING: IEEEtran.bst: No hyphenation pattern has been}%
\typeout{** loaded for the language `#1'. Using the pattern for}%
\typeout{** the default language instead.}%
\else
\language=\csname l@#1\endcsname
\fi
#2}}
\providecommand{\BIBdecl}{\relax}
\BIBdecl

\bibitem{rudenko2019human}
A.~Rudenko, L.~Palmieri, M.~Herman, K.~M. Kitani, D.~M. Gavrila, and K.~O.
  Arras, ``Human motion trajectory prediction: A survey,'' \emph{arXiv preprint
  arXiv:1905.06113}, 2019.

\bibitem{pellegrini2009you}
S.~Pellegrini, A.~Ess, K.~Schindler, and L.~van Gool, ``You'll never walk
  alone: Modeling social behavior for multi-target tracking,'' in
  \emph{Proc.~of the {IEEE} Int.~Conf.~on Computer Vision (ICCV)}, 2009, pp.
  261--268.

\bibitem{majecka2009statistical}
B.~Majecka, ``Statistical models of pedestrian behaviour in the forum,''
  \emph{Master's thesis, School of Informatics, University of Edinburgh}, 2009.

\bibitem{robicquet2016learning}
A.~Robicquet, A.~Sadeghian, A.~Alahi, and S.~Savarese, ``Learning social
  etiquette: Human trajectory understanding in crowded scenes,'' in
  \emph{Proc.~of the Europ. Conf.~on Comp. Vision (ECCV)}.\hskip 1em plus 0.5em
  minus 0.4em\relax Springer, 2016, pp. 549--565.

\bibitem{brscic2013person}
D.~Br\v{s}\v{c}i\'{c}, T.~Kanda, T.~Ikeda, and T.~Miyashita, ``Person tracking
  in large public spaces using 3-d range sensors,'' \emph{{IEEE} Trans. on
  Human-Machine Systems}, vol.~43, no.~6, pp. 522--534, 2013.

\bibitem{yan2017online}
Z.~Yan, T.~Duckett, and N.~Bellotto, ``Online learning for human classification
  in 3{D} {LiDAR}-based tracking,'' in \emph{Proc.~of the {IEEE} Int.~Conf.~on
  Intell. Robots and Syst. (IROS)}, 2017, pp. 864--871.

\bibitem{zhou2012understanding}
B.~Zhou, X.~Wang, and X.~Tang, ``Understanding collective crowd behaviors:
  Learning a mixture model of dynamic pedestrian-agents,'' in \emph{2012 IEEE
  Conference on Computer Vision and Pattern Recognition}.\hskip 1em plus 0.5em
  minus 0.4em\relax IEEE, 2012, pp. 2871--2878.

\bibitem{lerner2007crowds}
A.~Lerner, Y.~Chrysanthou, and D.~Lischinski, ``Crowds by example,'' in
  \emph{Computer Graphics Forum}, vol.~26, no.~3.\hskip 1em plus 0.5em minus
  0.4em\relax Wiley Online Library, 2007, pp. 655--664.

\bibitem{oh2011large}
S.~Oh \emph{et~al.}, ``A large-scale benchmark dataset for event recognition in
  surveillance video,'' in \emph{Proc.~of the {IEEE} Conf.~on Comp. Vis. and
  Pat. Rec. (CVPR)}, 2011, pp. 3153--3160.

\bibitem{Geiger2012CVPR}
A.~Geiger, P.~Lenz, and R.~Urtasun, ``Are we ready for autonomous driving? the
  kitti vision benchmark suite,'' in \emph{Conference on Computer Vision and
  Pattern Recognition (CVPR)}, 2012.

\bibitem{benfold2011stable}
B.~Benfold and I.~Reid, ``Stable multi-target tracking in real-time
  surveillance video,'' in \emph{Proc.~of the {IEEE} Conf.~on Comp. Vis. and
  Pat. Rec. (CVPR)}, 2011, pp. 3457--3464.

\bibitem{dondrup2015tracker}
C.~Dondrup, N.~Bellotto, F.~Jovan, and M.~Hanheide, ``Real-time multisensor
  people tracking for human-robot spatial interaction,'' in \emph{Proc.~of the
  {IEEE} Int.~Conf.~on Robotics and Automation (ICRA), Works. on ML for Social
  Robo.}\hskip 1em plus 0.5em minus 0.4em\relax IEEE, 2015.

\bibitem{lau2009tracking}
B.~Lau, K.~O. Arras, and W.~Burgard, ``Tracking groups of people with a
  multi-model hypothesis tracker,'' in \emph{Proc.~of the {IEEE} Int.~Conf.~on
  Robotics and Automation (ICRA)}, 2009.

\bibitem{linder2016multi}
T.~Linder, S.~Breuers, B.~Leibe, and K.~O. Arras, ``On multi-modal people
  tracking from mobile platforms in very crowded and dynamic environments,'' in
  \emph{Proc.~of the {IEEE} Int.~Conf.~on Robotics and Automation (ICRA)},
  2016.

\bibitem{Foka2010}
A.~F. Foka and P.~E. Trahanias, ``Probabilistic autonomous robot navigation in
  dynamic environments with human motion prediction,'' \emph{Int.~Journal of
  Social Robotics}, vol.~2, no.~1, pp. 79--94, 2010.

\bibitem{Bai2015}
H.~Bai, S.~Cai, N.~Ye, D.~Hsu, and W.~S. Lee, ``Intention-aware online pomdp
  planning for autonomous driving in a crowd,'' in \emph{Proc.~of the {IEEE}
  Int.~Conf.~on Robotics and Automation (ICRA)}, May 2015, pp. 454--460.

\bibitem{palmieri2017kinodynamic}
L.~Palmieri, T.~P. Kucner, M.~Magnusson, A.~J. Lilienthal, and K.~O. Arras,
  ``Kinodynamic motion planning on gaussian mixture fields,'' in \emph{2017
  IEEE International Conference on Robotics and Automation (ICRA)}.\hskip 1em
  plus 0.5em minus 0.4em\relax IEEE, 2017, pp. 6176--6181.

\bibitem{swaminathan2018down}
C.~S. Swaminathan, T.~P. Kucner, M.~Magnusson, L.~Palmieri, and A.~J.
  Lilienthal, ``Down the cliff: Flow-aware trajectory planning under motion
  pattern uncertainty,'' in \emph{2018 IEEE/RSJ International Conference on
  Intelligent Robots and Systems (IROS)}.\hskip 1em plus 0.5em minus
  0.4em\relax IEEE, 2018, pp. 7403--7409.

\bibitem{okal2016learning}
B.~Okal and K.~O. Arras, ``Learning socially normative robot navigation
  behaviors with bayesian inverse reinforcement learning,'' in \emph{Proc.~of
  the {IEEE} Int.~Conf.~on Robotics and Automation (ICRA)}, 2016.

\bibitem{chung2012incremental}
S.-Y. Chung and H.-P. Huang, ``Incremental learning of human social behaviors
  with feature-based spatial effects,'' in \emph{Proc.~of the {IEEE}
  Int.~Conf.~on Intell. Robots and Syst. (IROS)}, 2012, pp. 2417--2422.

\bibitem{Rudenko2018iros}
A.~Rudenko, L.~Palmieri, A.~J. Lilienthal, and K.~O. Arras, ``Human motion
  prediction under social grouping constraints,'' in \emph{Proc.~of the {IEEE}
  Int.~Conf.~on Intell. Robots and Syst. (IROS)}, 2018.

\bibitem{lasota2017survey}
P.~A. Lasota, T.~Fong, and J.~A. Shah, ``A survey of methods for safe
  human-robot interaction,'' \emph{Foundations and Trends in Robotics}, vol.~5,
  no.~4, pp. 261--349, 2017.

\bibitem{alahi2016social}
A.~Alahi, K.~Goel, V.~Ramanathan, A.~Robicquet, L.~Fei-Fei, and S.~Savarese,
  ``Social {LSTM}: Human trajectory prediction in crowded spaces,'' in
  \emph{Proc.~of the {IEEE} Conf.~on Comp. Vis. and Pat. Rec. (CVPR)}, 2016,
  pp. 961--971.

\bibitem{lo2019robust}
S.-Y. Lo, S.~Alkoby, and P.~Stone, ``Robust motion planning and safety
  benchmarking in human workspaces.'' in \emph{SafeAI@ AAAI}, 2019.

\bibitem{unhelkar2015human}
V.~V. Unhelkar, C.~P{\'e}rez-D'Arpino, L.~Stirling, and J.~A. Shah,
  ``Human-robot co-navigation using anticipatory indicators of human walking
  motion,'' in \emph{Proc.~of the {IEEE} Int.~Conf.~on Robotics and Automation
  (ICRA)}, 2015, pp. 6183--6190.

\bibitem{bera2017aggressive}
A.~Bera, T.~Randhavane, and D.~Manocha, ``Aggressive, tense, or shy?
  {I}dentifying personality traits from crowd videos,'' in \emph{Proc.~of the
  Int.~Conf.~on Artificial Intelligence (IJCAI)}, 2017, pp. 112--118.

\bibitem{ma2016forecasting}
W.-C. Ma, D.-A. Huang, N.~Lee, and K.~M. Kitani, ``Forecasting interactive
  dynamics of pedestrians with fictitious play,'' in \emph{Proc.~of the {IEEE}
  Conf.~on Comp. Vis. and Pat. Rec. (CVPR)}, 2017, pp. 4636--4644.

\bibitem{molina2018modelling}
S.~Molina, G.~Cielniak, T.~Krajn{\'\i}k, and T.~Duckett, ``Modelling and
  predicting rhythmic flow patterns in dynamic environments,'' in \emph{Annual
  Conf. Towards Autonom. Rob. Syst.}\hskip 1em plus 0.5em minus 0.4em\relax
  Springer, 2018, pp. 135--146.

\bibitem{kucner2017enabling}
T.~P. Kucner, M.~Magnusson, E.~Schaffernicht, V.~H. Bennetts, and A.~J.
  Lilienthal, ``Enabling flow awareness for mobile robots in partially
  observable environments,'' \emph{{IEEE} Robotics and Automation Letters},
  vol.~2, no.~2, pp. 1093--1100, 2017.

\bibitem{doshi2009roles}
A.~Doshi and M.~M. Trivedi, ``On the roles of eye gaze and head dynamics in
  predicting driver's intent to change lanes,'' \emph{IEEE Transactions on
  Intelligent Transportation Systems}, vol.~10, no.~3, pp. 453--462, 2009.

\bibitem{palinko2016robot}
O.~Palinko, F.~Rea, G.~Sandini, and A.~Sciutti, ``Robot reading human gaze: Why
  eye tracking is better than head tracking for human-robot collaboration,'' in
  \emph{Proc.~of the {IEEE} Int.~Conf.~on Intell. Robots and Syst. (IROS)},
  2016.

\bibitem{admoni2017social}
H.~Admoni and B.~Scassellati, ``Social eye gaze in human-robot interaction: a
  review,'' \emph{Journal of Human-Robot Interaction}, vol.~6, no.~1, pp.
  25--63, 2017.

\bibitem{kiefer2019eye}
P.~Kiefer, I.~Giannopoulos, M.~Raubal, and A.~Duchowski, ``Eye tracking for
  spatial research: Cognition, computation, challenges,'' \emph{Spatial
  Cognition \& Computation}, vol.~17, no. 1-2, pp. 1--19, 2017.

\bibitem{chadalavada2020bi}
R.~T. Chadalavada, H.~Andreasson, M.~Schindler, R.~Palm, and A.~J. Lilienthal,
  ``Bi-directional navigation intent communication using spatial augmented
  reality and eye-tracking glasses for improved safety in human--robot
  interaction,'' \emph{Robotics and Computer-Integrated Manufacturing},
  vol.~61, p. 101830, 2020.

\bibitem{moussaid2010walking}
M.~Moussa{\"\i}d, N.~Perozo, S.~Garnier, D.~Helbing, and G.~Theraulaz, ``The
  walking behaviour of pedestrian social groups and its impact on crowd
  dynamics,'' \emph{PloS one}, vol.~5, no.~4, p. e10047, 2010.

\end{thebibliography}
}

\end{document}